\documentclass[review]{elsarticle}

\usepackage[english]{babel}
\usepackage[T1]{fontenc}
\usepackage{etoolbox}
\makeatletter

\def\ps@pprintTitle{%
   \let\@oddhead\@empty
   \let\@evenhead\@empty
   \let\@oddfoot\@empty
   \let\@evenfoot\@oddfoot
}
\makeatother

\usepackage{subfig}
\usepackage{float}
\usepackage{caption}

\usepackage{comment}
\usepackage{tabularx}
\usepackage{subfig}
\usepackage{wrapfig}
\usepackage{hyperref}
\hypersetup{pdfauthor={Conor}}
\usepackage{float}
\usepackage{amsmath}

\usepackage{amsmath, amssymb, amsfonts}
\usepackage{algorithmic}
\usepackage{graphicx}
\graphicspath{{figures}}
\usepackage{url}
\usepackage{lipsum}
\usepackage{balance}
\usepackage{textcomp}
\usepackage{xcolor}
\usepackage{tabto}
\usepackage{verbatim}
\usepackage{multirow}
\usepackage{array}
\usepackage{color}
\usepackage{bm}
\usepackage{lscape}
\usepackage{nomencl}
\makenomenclature
\usepackage{tikz}
\usepackage{etoolbox}
\renewcommand\nomgroup[1]{%
  \item[\bfseries
  \ifstrequal{#1}{A}{In this paper we have used the following abbreviations for the algorithms used.}{}%
]}

\usepackage[switch]{lineno}

\begin{document}
\begin{frontmatter}

\title{Enhancing Coastal Water Body Segmentation with Landsat Irish Coastal Segmentation (LICS) Dataset}

\author[add1,add2]{Conor~O'Sullivan}
\ead{conor.osullivan4@ucdconnect.ie}
\author[add3]{Ambrish~Kashyap}
\ead{Ambrish.Kashyap.ch719@chemical.iitd.ac.in}
\author[add4]{Seamus~Coveney}
\author[add5]{Xavier~Monteys}
\author[add1,add2]{Soumyabrata Dev\corref{mycorrespondingauthor}}
\cortext[mycorrespondingauthor]{Corresponding author. Tel.: + 353 1896 1797.}
\ead{soumyabrata.dev@ucd.ie}

\address[add1]{ADAPT SFI Research Centre, Dublin, Ireland }
\address[add2]{School of Computer Science, University College Dublin, Ireland}
\address[add3]{Indian Institute of Technology Delhi, India}
\address[add4]{Envo-Geo Environmental Geoinformatics, Skibbereen, Ireland}
\address[add5]{Geological Survey Ireland, Dublin, Ireland}

\begin{abstract}
Ireland's coastline, a critical and dynamic resource, is facing challenges such as erosion, sedimentation, and human activities. Monitoring these changes is a complex task we approach using a combination of satellite imagery and deep learning methods. However, limited research exists in this area, particularly for Ireland. This paper presents the Landsat Irish Coastal Segmentation (LICS) dataset, which aims to facilitate the development of deep learning methods for coastal water body segmentation while addressing modelling challenges specific to Irish meteorology and coastal types. The dataset is used to evaluate various automated approaches for segmentation, with U-NET achieving the highest accuracy of 95.0\% among deep learning methods. Nevertheless, the Normalized Difference Water Index (NDWI) benchmark outperformed U-NET with an average accuracy of 97.2\%. The study suggests that deep learning approaches can be further improved with more accurate training data and by considering alternative measurements of erosion. The LICS dataset and code are freely available to support reproducible research and further advancements in coastal monitoring efforts.
\end{abstract}


\end{frontmatter}

\section{Introduction} 
\label{sec:intro}
Ireland's coastline is both a vital and dynamic resource. Coastal regions are impacted by erosion, sedimentation, and human activities like land development. In fact, it is estimated that 20\% of Ireland's 4,578km of coastline are eroding~\cite{eurosion2004}. A trend that is likely to be exacerbated by climate change and sea-level rise~\cite{masselink2013impacts}. To identify the areas worst at risk we must closely monitor changes in the coastline. 

The length of the coastline means this is no straightforward task. There is a growing consensus that, to meet the challenge, we can use a combination of satellite imagery and deep learning methods~\cite{seale2022swed}. At the same time, there is limited research done in this area. Particularly for Ireland, there are no extensive open-source machine-learning datasets for coastal water body segmentation.

Hence, we present the Landsat Irish Coastal Segmentation (LICS) dataset. Its purpose is to aid the development of deep learning methods for coastal water body segmentation. At the same time, the dataset may be used to shed light on modelling challenges specific to Ireland. In particular, we aim to answer questions about how solar altitude, various coastline types and the date of images will impact model performance. In the process, we benchmark various automated approaches for segmentation and explore their assumptions. In the spirit of reproducible research, both the dataset\footnote{The LICS dataset can be found here: \url{https://doi.org/10.5281/zenodo.8414665}}  and code\footnote{The code used to produce all results can be found here: \url{https://github.com/conorosully/landsat-coastline-segmentation}} are freely available.

\section{Background}
\label{sec:background}

We must distinguish between two tasks -- coastal water body segmentation and coastline detection. For segmentation, we aim to classify each pixel in an image as either land or ocean. For coastline detection, we aim to classify each pixel as either coastline or not. The latter process will depend on how we define the coastline. In this paper, we consider the instantaneous coastline which is the boundary between land and water at the exact time a satellite image was taken~\cite{sun2023coastline}. Under this definition, the two tasks are related. That is the coastline pixels are the pixels where the segmentation map changes from land to ocean. 

The instantaneous coastline is limited in its ability to measure erosion as it depends on the tide. Alternative measurements include the high water mark, vegetation line and dune volume~\cite{hanslow2007beach}. These are considered to be better definitions for measuring erosion. However, gathering ground truth for these measurements is more complicated as they require onsite evaluation. In comparison, the instantaneous coastline can be determined using only satellite images and additional higher-resolution images of the same coastline~\cite{Xiong2023, seale2022swed}. This partly explains why most studies have chosen this definition and approach to creating a ground truth dataset.

Traditionally, spectral indices have been used for water body segmentation~\cite{liu2022comparison, osullivan2023analyzing}. For coastline detection, various edge detection algorithms have been applied~\cite{vukadinov2017algorithm, klinger2011antarctic, paravolidakis2018automatic}. The advantage of these approaches is they do not require a training set.  The downside is they are not robust to noise in satellite images caused by factors like clouds, swell and land development~\cite{osullivan2023igrass,wu2023measurement,mcnicholl2021evaluating}. Satellite images and ground-based sky images~\cite{jain2021using} are often corrupted by atmospheric clouds~\cite{dev2016machine,dev2019multi,dev2018high}. Additionally, as they require one channel as input, we must first select~\cite{dev2016rough} an individual spectral band or combine multiple bands into one value per pixel. In the process, we may lose important information from other bands or from interactions between bands. 

In comparison, deep learning models can use all available spectral bands. Additionally, they can use a pixel's context to make predictions. This means they can use the spectral band intensities from surrounding pixels and not just the intensities for the given pixel. Initial work with these models has shown promise. ~\cite{li2018deepunet},~\cite{cheng2016senet} and~\cite{Shamsolmoali2019} apply variations of U-NET, a common image segmentation algorithm, to coastal water body segmentation datasets.  However, the images in the studies are naturally coloured meaning the models cannot make use of the range of spectral bands available in satellite images. Particularly, the Near-Infrared (NIR) band which is important for water body segmentation~\cite{mondejar2019near, osullivan2023interpreting}. 

To the best of our knowledge, four studies use satellite images as input.~\cite{Vos2019} showed a multi-layer perception could accurately segment five coastal water bodies across three continents.~\cite{rogers2021vedgedetector} focused on predicting the vegetation line using convolutional neural networks (CNN).~\cite{Xiong2023} used a combination of CNN and transformer architecture for land-sea segmentation in the yellow sea region of china. In terms of dataset diversity,~\cite{seale2022swed} presents the most extensive study. The researchers provided a test set of 98 images from 49 locations around the world. The aim was to provide a benchmark dataset that would aid the development of land-ocean segmentation models that are scalable to all global coastlines. 

Such a model is ideal. However, it is a challenging task. All coastal regions will have their own unique geographical and meteorological conditions~\cite{manandhar2018data,wu2022linkclimate,jain2020forecasting} and labelling a training dataset that adequately captures these variations will be time-consuming. Hence, ~\cite{seale2022swed} opted to use semi-supervised methods to label their training dataset. To make the task more manageable, we have chosen to focus on one country---Ireland. Still, even this relatively small island presents a large variation in coastline conditions. 

From sandy beaches to rocky cliffs, Ireland's varying coastal geographies will make some coastlines more or less susceptible to erosion~\cite{gault2011erosion, thebaudeau2013modelling}. Wave power is another factor that affects erosion~\cite{smyth2023nearshore}. The west coast of Ireland faces the Atlantic and experiences a larger amount of wave energy~\cite{o2020updated}. The long-term effect is typically more jagged coastlines in these areas. In other words, we have a less uniform boundary between land and ocean and we expect these areas to be more challenging to produce accurate segmentation. 

Other considerations are cloud cover, tidal variations and variations in solar altitude---the angle of elevation of the sun above the horizontal plane. Ireland experiences large differences in solar altitude between summer and winter months. A factor worth considering as low solar altitudes have been shown to lead to poorer performance for water body extraction indices~\cite{kaplan2017water}. Ultimately, if we want a model that can perform accurate segmentation across all times and coastline types, we must build a dataset that adequately captures variation in these factors. 

\section{Methodology}
\label{sec:methodology}

\subsection{Landsat Irish Coastal Segmentation Dataset}
\begin{figure}[ht]
\centering
\includegraphics[width=0.80\textwidth]{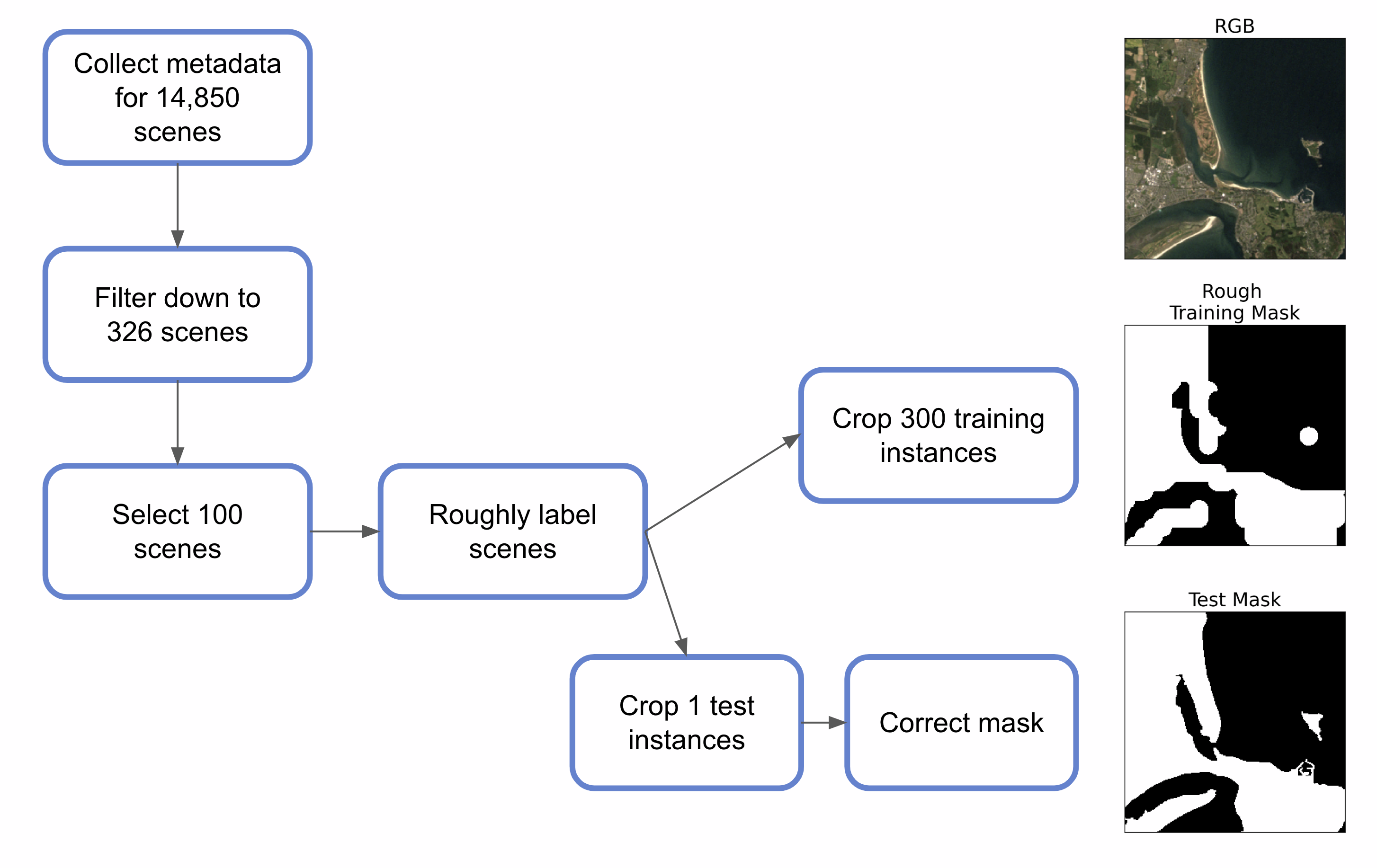}
\caption{Summary of the Landsat scene selection, scene cropping and annotation process. The end result of the process is 30,000 training instances and 100 test instances.}
\label{fig:dataset_outline}
\end{figure}

We introduce the Landsat Irish Coastal Segmentation (LICS) dataset~\cite{lics2023}. This is the first dataset created for deep-learning semantic segmentation of the Irish coastline. It has been created with the goal of developing robust models that can perform accurate segmentation across different years, coastal types and atmospheric conditions. Particular attention has been paid to the model performance at varying solar altitudes. Figure \ref{fig:dataset_outline} gives a summary of the dataset development process and we will discuss each step in depth. 

\subsubsection*{Selecting Scenes}
\begin{figure}[ht]
\centering
\includegraphics[width=0.98\textwidth]{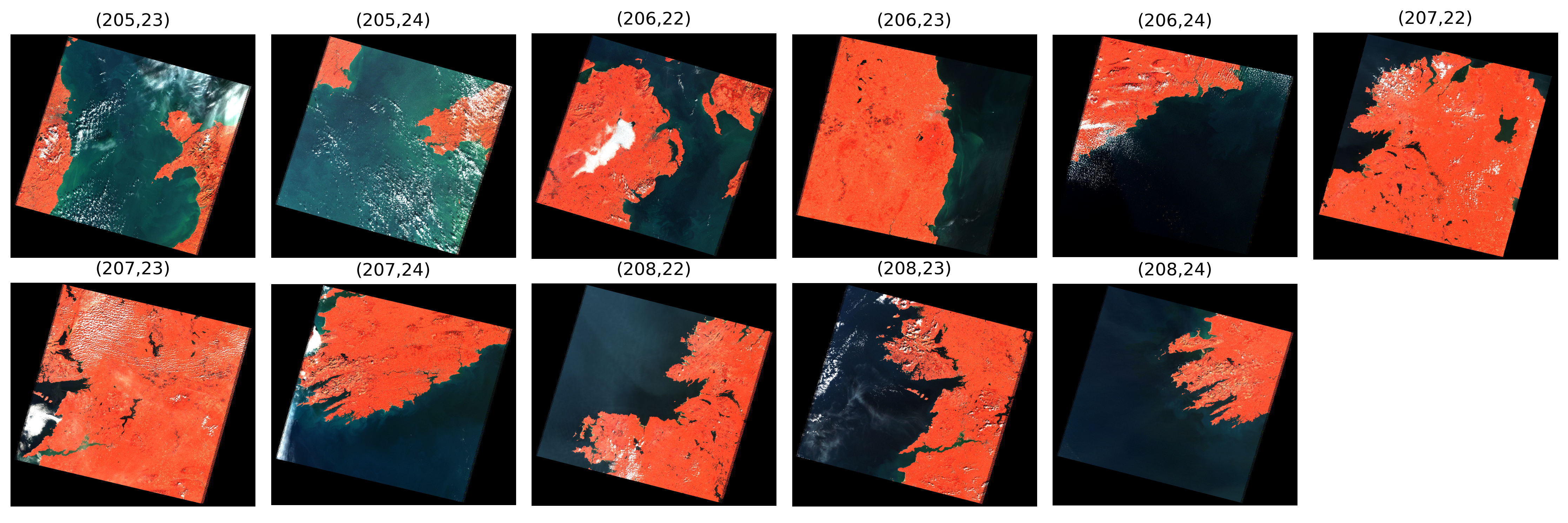}
\caption{An example of each of the 11 Landsat tiles considered for this analysis. The tile's row and path (row,path) are given in the title above each image. The scenes have been visualised using the NIR band to show contrast between land and ocean. }
\label{fig:tiles}
\end{figure}

The first step was to obtain metadata of all potential Landsat scenes. A tile covers a specific geographic area and we considered 11 tiles which all contained some section of the Irish coastline. You can see examples of these in Figure \ref{fig:tiles}. Combined, every section of the Irish coastline is included in these 11 tiles. We obtained the metadata for all scenes from these tiles from April 1984 to May 2023. This was 14,850 scenes in total. 

The fields included in the metadata allowed us to select scenes from this list for model development. Specifically, we removed any scenes that did not meet the following criteria:
\begin{enumerate}
    \item We select scenes from Landsat 5, 7, 8 \& 9.
    \item For Landsat 7, we only select scenes before 2003-05-31 due to faulty satellite mirrors after this date. 
    \item We select scenes that fall in Tier 1 as these are the highest quality data. 
    \item We consider scenes that had less than 10\% total cloud cover.
\end{enumerate}

\begin{figure}[ht]
\centering
\includegraphics[width=0.80\textwidth]{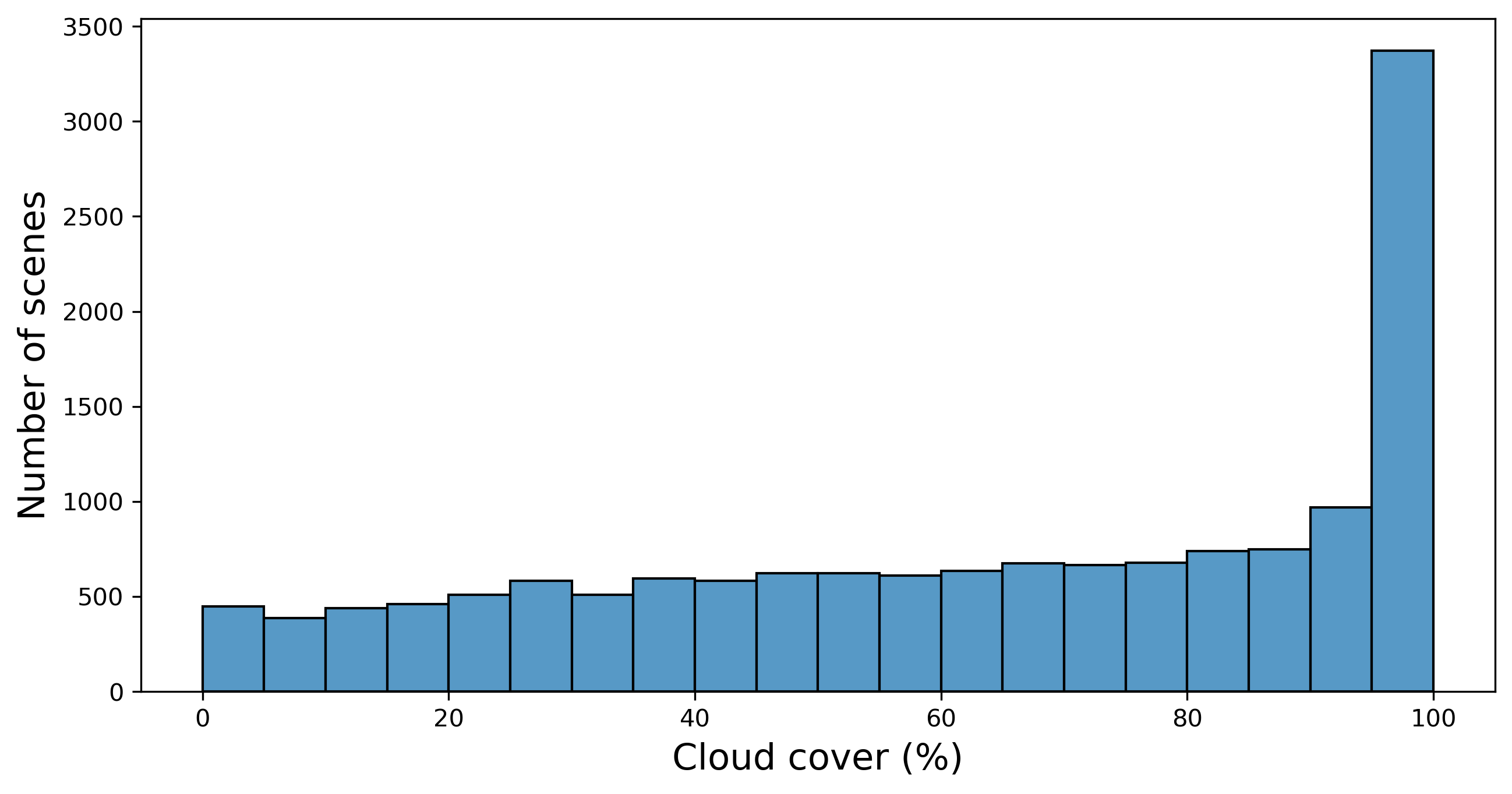}
\caption{Frequency of cloud cover percentage. The frequencies are calculated using the metadata of 14,850 Landsat scenes of Ireland.}
\label{fig:cloud}
\end{figure}

The cloud cover percentage is calculated using the CFMask algorithm~\cite{cfmask2017}. Figure~\ref{fig:cloud} gives the histogram of these cloud cover percentages for all the scenes. Ideally, we would only select scenes that had 0\% cloud cover. However, we can see that this would severely limit the number of available scenes. In fact, only 5.6\% of scenes had less than 10\% cloud cover so we decided to use this as our cutoff. 

\begin{figure}[ht]
\centering
\includegraphics[width=0.80\textwidth]{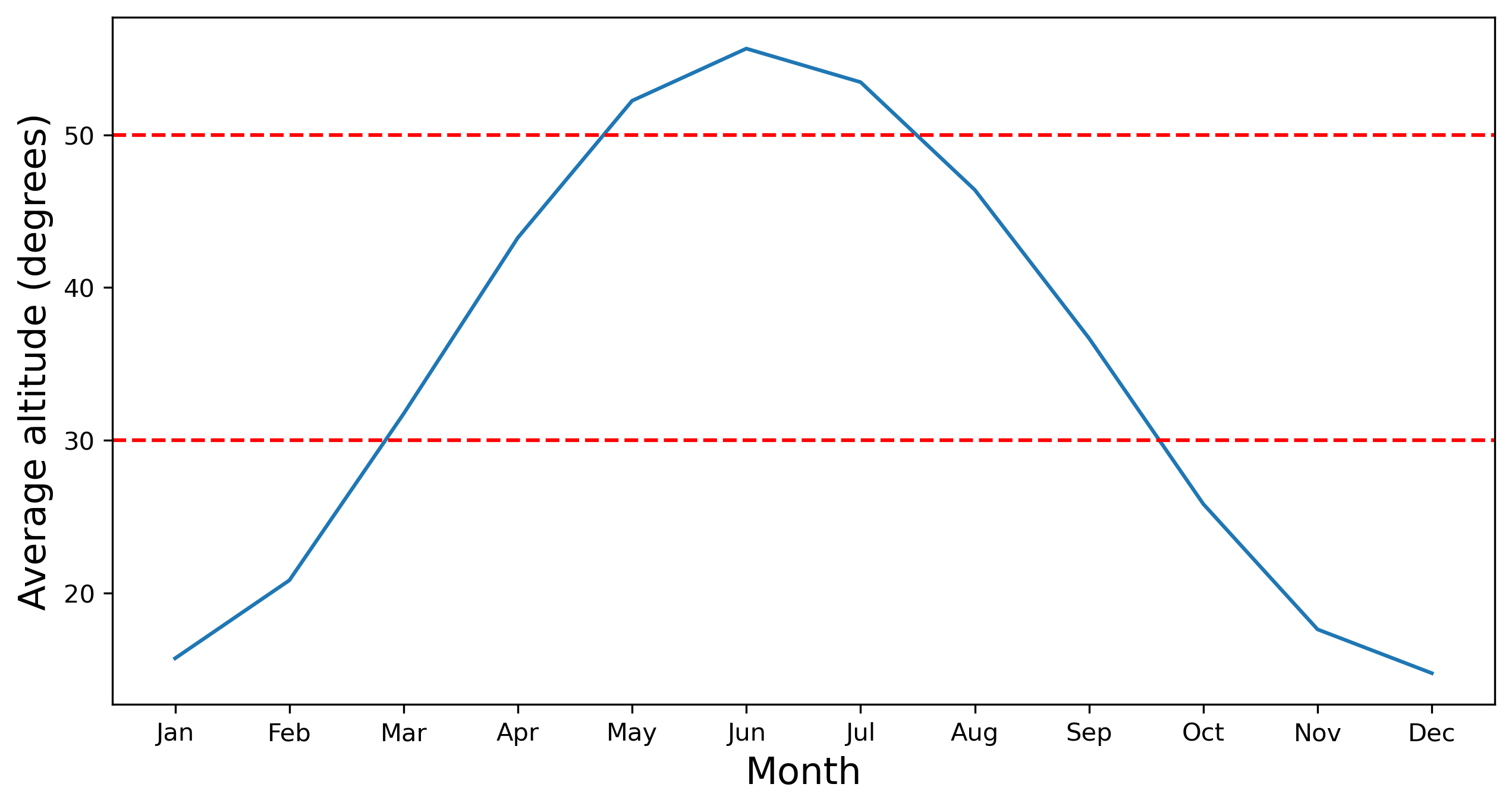}
\caption{Average solar altitude by month of 14,850 Landsat scenes of Ireland. We take the altitude of the sun at the location and time the scenes were taken. We can see that the altitude is highest in the summer months.}
\label{fig:altitude}
\end{figure}

The above process left us with 326 scenes. We selected 100 scenes from this list using the solar altitude as an additional criterion. We calculated the altitude using the time and geolocation of a scene. The average altitude by month is given in Figure~\ref{fig:altitude}. As shown by the red lines, we divided the scenes into high ($>50$ degrees), medium ($>30$ degrees) and low ($<=30$ degrees) altitude categories. These groupings were chosen as they divided the scenes evenly into three groups.

For each year and altitude category, we selected the scene that had the lowest cloud cover. In Table \ref{tab:tile_count} we see the breakdown of scenes for each tile. To have a more even distribution across the tiles, we selected a further 8 scenes for tile 6. The final dataset had 42, 42 and 43 scenes in the altitude categories respectively and at least one scene in each year. The final result is a dataset that captures variation introduced by solar altitude, coastline type and time. As altitude is related to the time of year, we also capture month-on-month variation. 

\begin{table}[ht]
\caption{The number of scenes selected for each tile.}
\centering
\label{tab:tile_count}
\begin{tabular}{ccc|cc|}
\hline
\multicolumn{1}{|c|}{\textbf{Tile}} & \multicolumn{1}{c|}{\textbf{Path}} & \textbf{Row}          & \multicolumn{1}{c|}{\textbf{\begin{tabular}[c]{@{}c@{}}Initial \\ selection\end{tabular}}} & \textbf{\begin{tabular}[c]{@{}c@{}}Additional\\ selection\end{tabular}} \\ \hline
\multicolumn{1}{|c|}{1}             & \multicolumn{1}{c|}{205}           & 23                    & \multicolumn{1}{c|}{11}                                                                    & 0                                                                       \\ \hline
\multicolumn{1}{|c|}{2}             & \multicolumn{1}{c|}{205}           & 24                    & \multicolumn{1}{c|}{20}                                                                    & 0                                                                       \\ \hline
\multicolumn{1}{|c|}{3}             & \multicolumn{1}{c|}{206}           & 22                    & \multicolumn{1}{c|}{9}                                                                     & 0                                                                       \\ \hline
\multicolumn{1}{|c|}{4}             & \multicolumn{1}{c|}{206}           & 23                    & \multicolumn{1}{c|}{6}                                                                     & 0                                                                       \\ \hline
\multicolumn{1}{|c|}{5}             & \multicolumn{1}{c|}{206}           & 24                    & \multicolumn{1}{c|}{10}                                                                    & 0                                                                       \\ \hline
\multicolumn{1}{|c|}{6}             & \multicolumn{1}{c|}{207}           & 22                    & \multicolumn{1}{c|}{1}                                                                     & 8                                                                       \\ \hline
\multicolumn{1}{|c|}{7}             & \multicolumn{1}{c|}{207}           & 23                    & \multicolumn{1}{c|}{10}                                                                    & 0                                                                       \\ \hline
\multicolumn{1}{|c|}{8}             & \multicolumn{1}{c|}{207}           & 24                    & \multicolumn{1}{c|}{7}                                                                     & 0                                                                       \\ \hline
\multicolumn{1}{|c|}{9}             & \multicolumn{1}{c|}{208}           & 22                    & \multicolumn{1}{c|}{6}                                                                     & 0                                                                       \\ \hline
\multicolumn{1}{|c|}{10}            & \multicolumn{1}{c|}{208}           & 23                    & \multicolumn{1}{c|}{6}                                                                     & 0                                                                       \\ \hline
\multicolumn{1}{|c|}{11}            & \multicolumn{1}{c|}{208}           & 24                    & \multicolumn{1}{c|}{6}                                                                     & 0                                                                       \\ \hline
\multicolumn{1}{l}{}                & \multicolumn{1}{l}{}               & \multicolumn{1}{l|}{} & \multicolumn{2}{c|}{100}                                                                                                                                             \\ \cline{4-5} 
\end{tabular}
\end{table}
\subsubsection*{Spectral bands}

After selecting the final list, we obtained the spectral bands for the 100 scenes. We consider the bands listed in Table \ref{tab:bands} as input into the modelling approaches. These all have a resolution of 30m. These are the bands common to Landsat 5, 7, 8 and 9. The newer satellites do have more bands available. However, we believe the ones we have selected to be appropriate for water body segmentation as they include bands common to water body indices.

\begin{table}[h]
\caption{The spectral bands used as input into segmentation approaches. They all have a resolution of 30m.}
\centering
\label{tab:bands}
\begin{tabular}{c|c|c|}
\cline{2-3}
                        & \textbf{Band}        & \textbf{Accronym} \\ \hline
\multicolumn{1}{|c|}{1} & Blue                 & B                 \\ \hline
\multicolumn{1}{|c|}{2} & Green                & G                 \\ \hline
\multicolumn{1}{|c|}{3} & Red                  & R                 \\ \hline
\multicolumn{1}{|c|}{4} & Near Infrared        & NIR               \\ \hline
\multicolumn{1}{|c|}{5} & Shortwave Infrared 1 & SWIR1             \\ \hline
\multicolumn{1}{|c|}{6} & Shortwave Infrared 2 & SWIR2             \\ \hline
\multicolumn{1}{|c|}{7} & Thermal              & T                 \\ \hline
\end{tabular}
\end{table}

\subsubsection*{Cropping scenes}
\begin{figure}[ht]
\centering
\includegraphics[width=0.80\textwidth]{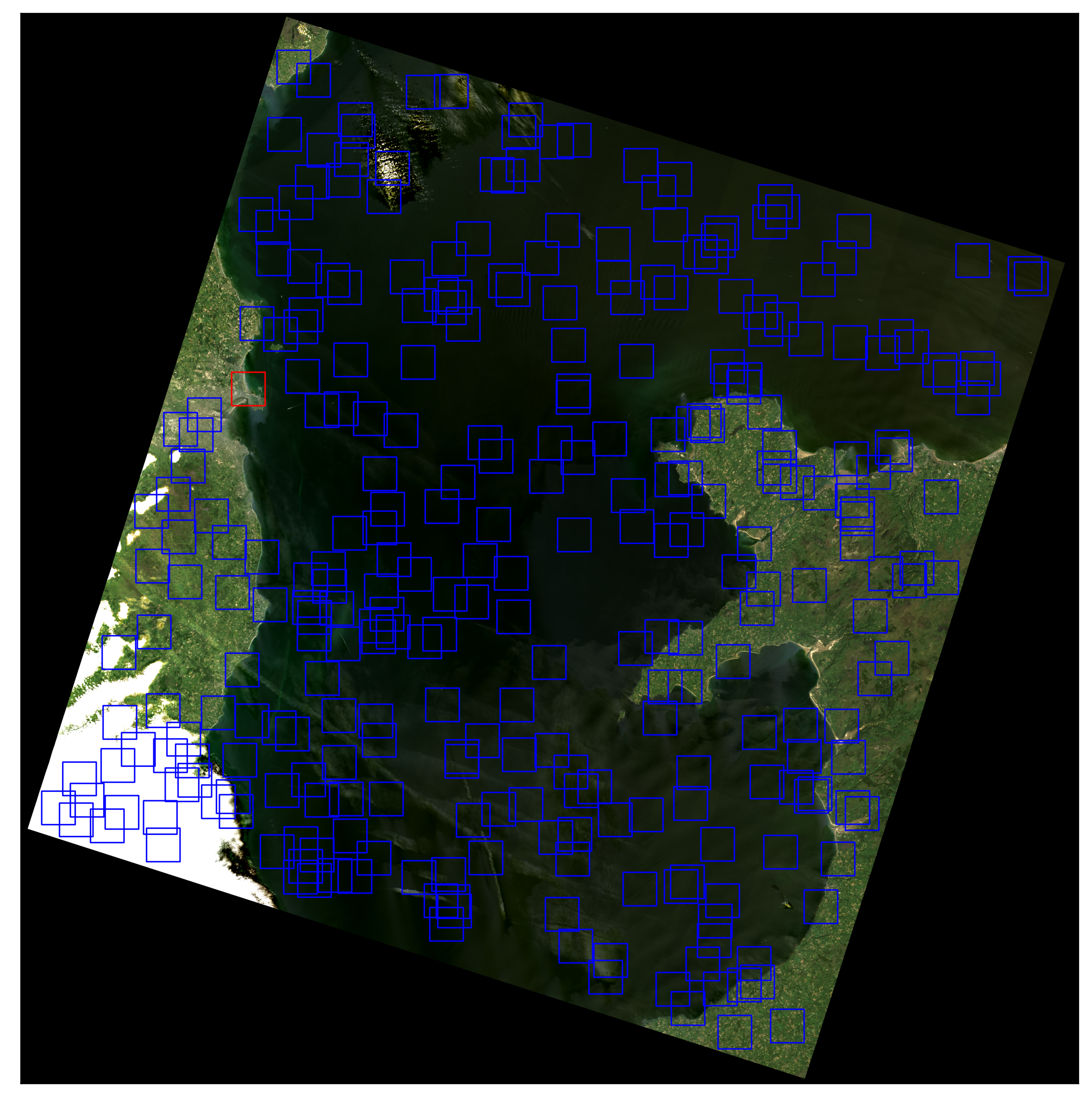}
\caption{Example of test and training crops from a Landsat scene with tile (205,23). The test crop is given by the red square. The training crops are shown by the 300 blue squares.}
\label{fig:crop}
\end{figure}

The Landsat scenes are roughly 8,000 by 8,000 pixels. These dimensions are larger than what is typically used to train machine learning models. Hence, as seen in Figure~\ref{fig:crop} we crop 256 by 256 pixels squares from each scene to create the training and test set. For the test set, we select one geographical location for each tile. Hence, we have 11 testing locations with additional variation introduced through time and atmospheric conditions. These locations are chosen randomly with the conditions that they fall on the island of Ireland, no bounding box is included and the ratio of land to ocean is between 40\% and 60\%. 

For the training set, 300 crops per scene are selected. These are chosen randomly with the condition that they do not overlap with the testing location and contain no bounding box. Each training instance was randomly flipped vertically with 50\% probability and horizontally with 50\% probability. The final result is a dataset of 30,000 training instances and 100 test instances. Importantly, the test set is geographically independent of the training set. Hence, evaluation results will indicate the model's ability to generalise to the Irish coastline. 

\subsubsection*{Training Annotation}
The training annotations were created manually by drawing segmentation masks on top of the Landsat scenes. Specifically, pixels were given a value of 1 for ocean and 0 otherwise. To be clear, a scene was annotated before the above cropping process and then the masks were cropped along with the spectral bands. This approach was chosen as it was less time-consuming than annotating the 30,000 training instances individually.  

\begin{figure}[ht]
\centering
\includegraphics[width=0.99\textwidth]{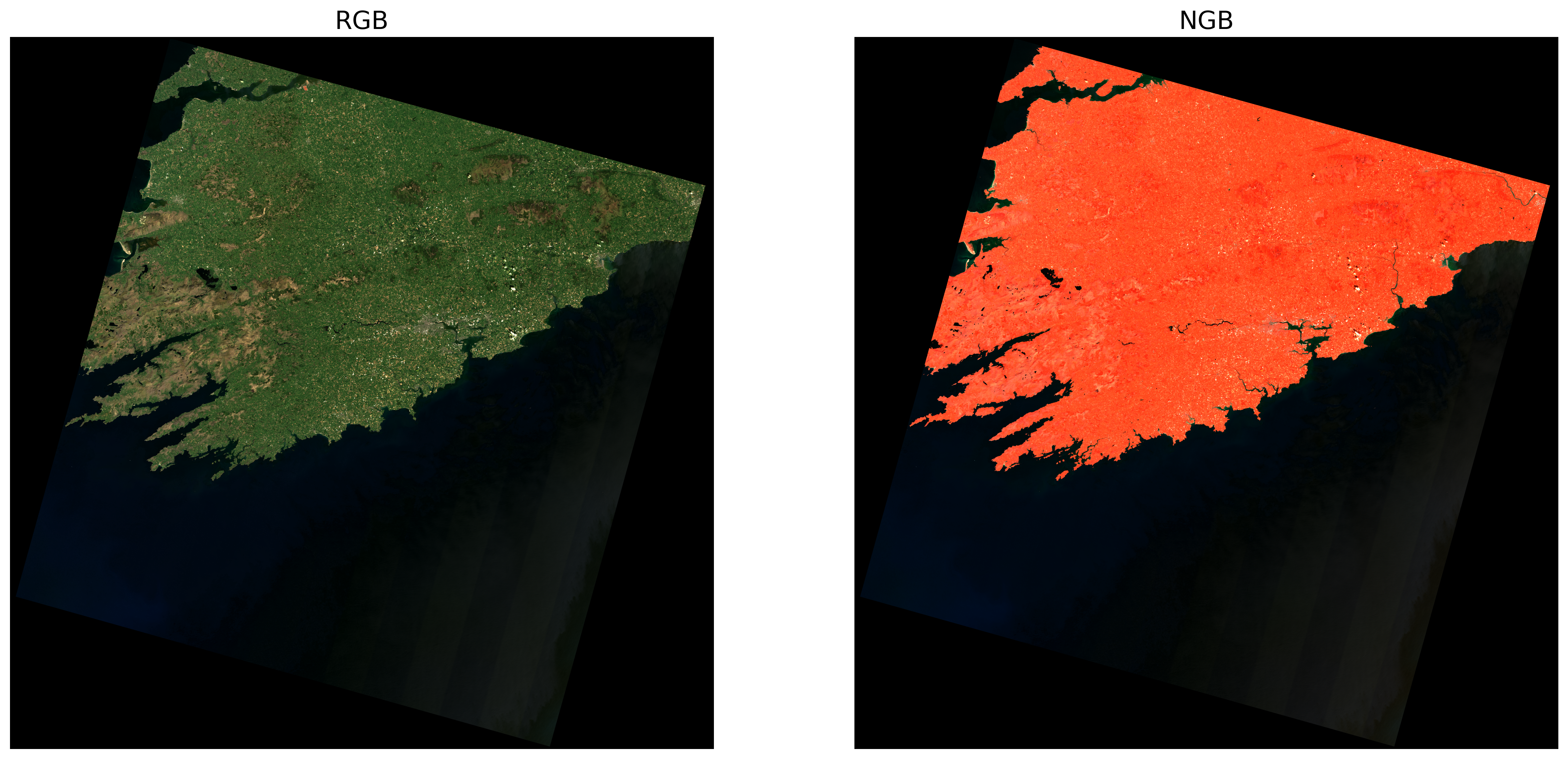}
\caption{Example of the references used to annotate the training set. These include a visualization of the visible light bands (RGB) and a visualisation which uses the Near-infrared band in place of the red band (NGB).}
\label{fig:training_ref}
\end{figure}

To further reduce time requirements only a rough mask was drawn. These typically took between 15 and 25 minutes depending on the tile and a strict cutoff of 30 minutes per scene was used. As a reference when drawing the masks, the scenes were visualised using the standard RGB (3/2/1) bands and using the NIR band in replace of the Red band (4/2/1). These can be seen in Figure~\ref{fig:training_ref}. Open-source software called Label Studio was used to draw the annotations.

\subsubsection*{Test Annotation}
\begin{figure}[ht]
\centering
\includegraphics[width=0.99\textwidth]{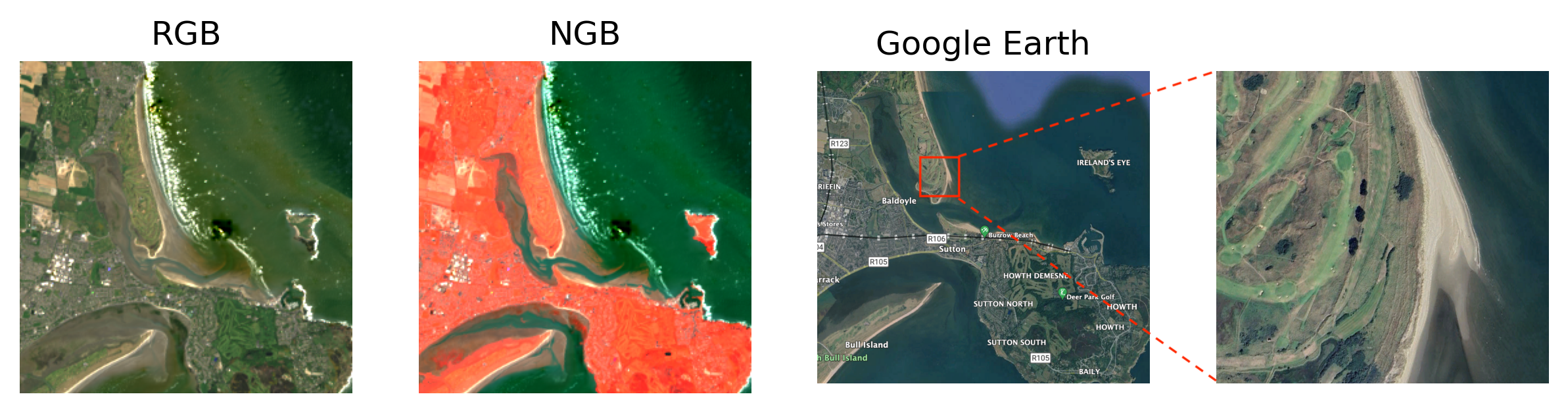}
\caption{Example of the references used to annotate the test set. These include a visualization of the visible light bands (RGB) and a visualisation which uses the Near-infrared band in place of the red band (NGB) and high-resolution Google Earth images from multiple time periods. }
\label{fig:labelling_ref}
\end{figure}

Evaluating models using these rough annotations would likely overestimate model performance. Hence, for the test instances, we created more precise annotations. This was done after the cropping process and so only the pixels within the test crop area were annotated. Figure~\ref{fig:labelling_ref} gives the references used to create the test annotations. Like the training set, these instances were visualised using the 3/2/1 and 4/2/3 bands. Additionally, Google Earth Pro was used to provide a higher-resolution image of the testing locations. This allowed us to observe the location at various tide levels and at times close to when the Landsat scene was taken. No time limit was set to ensure the most accurate annotations possible.

The problem of mixed pixels should be mentioned. These are land and ocean pixels in a Landsat scene that have merged. This effect will be most prominent in pixels close to the instantaneous coastline. As we have decided on a binary target variable, these must either be classified as land or ocean. In the test set, these are handled by the authors' judgement based on the available resources mentioned above. Overall, the process produced segmentation masks that reflected the true instantaneous coastline as closely as possible without visiting the testing locations. 

We can see an example of a rough training mask and a more precise test mask for a testing location in Figure~\ref{fig:dataset_outline}. The hope is the mistakes in the rough masks are not systematic. Then through training on 30,000 instances, the mistakes will be averaged out and we will be able to predict an accurate segmentation. Evaluating the segmentation approaches using the more precise test masks will give a clearer indication of the true performance of the models. However, the results should be interpreted with the test set annotation process in mind. 

\subsubsection{Coastal Type Classification}
For further analysis, the test images were classified by their coastline types --- "rocky" or "sandy". We consider only these classifications as it is estimated that the majority of Ireland’s coast is either hard rock (59\%) or sandy beaches (39\%)~\cite{eurosion2004}. All testing locations are classified visually using the same references seen in Figure~\ref{fig:labelling_ref}. For locations with mixed types, the majority type was used for the final classification. These include tiles (207,22), (208,22) and (208,24). They were classified as rocky but a minority of the coastline was sandy.   

\subsection{Segmentation Approaches}
\subsubsection*{Normalized Difference Water Index}
As a benchmark, we use the Normalized Difference Water Index (NDWI)~\cite{ndwi1996}. This is a well-established spectral index for water body extraction. As seen in Equation~\ref{eq:ndwi}, an intensity value is calculated for each pixel in a test image. If this value is equal to or above 0 the pixel is labelled as water. If it is less than 0 the pixel is labelled as land. As this process is deterministic, it does not require the training set.

\begin{equation}\label{eq:ndwi}
 NDWI = \frac{G - NIR}{G + NIR}
\end{equation}

\subsubsection*{Extreme Gradient Boosting}
For comparison to the deep learning methods, we used an Extreme Gradient Boosting (XGBoost) model~\cite{xgboost2016}. This is an ensemble method that makes predictions using a collection of decision trees. Specifically, we used a model with 500 trees and a maximum depth of 3 for each tree. To create the dataset for this model, we randomly select 100 pixels from each training image. This gives us 3,000,000 rows where each row has 8 values---one for each band and the target variable. After training, the model is used to classify each pixel in a test image individually. The predictions are then combined into the final segmentation prediction.

\subsubsection*{U-Net}
For the deep learning method, we use the U-Net architecture~\cite{unet2015}. This is a popular segmentation architecture developed for medical image segmentation. The architecture consists of an encoder, bottleneck and decoder. Layers in the encoder and decoder are connected through skip connections. The model was trained using a 90/10 training/evaluation split for 50 epochs with early stopping if the validation loss did not improve for 10 epochs. We follow the same process using variations of the U-Net. That is the Attention U-Net~\cite{attunet2018} and R2 U-Net~\cite{r2unet2018}. It is not clear if these variations will provide improvement in performance for our problem. This is because they have been developed to address issues common to medical imagery--- small sample sizes and unbalanced datasets.

As input, the 3 deep learning approaches take all bands and pixels. This means they can not only use the spectral bands for a pixel but also the surrounding pixels to make segmentation predictions. We expect this to improve model performance. Especially for pixels close to the coastline where we expect the distinction between land and ocean pixels to be less clear. By comparing these approaches to the XGBoost model we can understand the extent to which this is true. 

\subsection{Evaluation Metrics}

\subsubsection*{Confusion Matrix Metrics}
When evaluating the segmentation approaches we consider confusion matrix-based measures based on the values in Table~\ref{tab:confusion_matrix}. Suppose $P_{i,j}$ and $G_{i,j}$ are the pixel values in the $i,jth$ position in the predicted segmentation mask (P) and the ground truth mask (G). Then TP is the count of cases where $P_{i,j} = G_{i,j} = 1$, TN is the count where $P_{i,j} = G_{i,j} = 0$, FP is the count where $P_{i,j} = 1, G_{i,j} = 0 $ and FN is the count where $P_{i,j} = 0, G_{i,j} = 1$. We use the metrics based on these values listed in Equations~\ref{eq:accuracy}-~\ref{eq:f1}.

\begin{table}[ht]
\centering

\begin{tabular}{llcc}
                                                                                                       &                        & \multicolumn{2}{c}{\textbf{Prediction}}                                                                                                                                   \\ \cline{3-4} 
                                                                                                       & \multicolumn{1}{l|}{}  & \multicolumn{1}{c|}{1}                                                              & \multicolumn{1}{c|}{0}                                                              \\ \cline{2-4} 
\multicolumn{1}{c|}{\multirow{2}{*}{\textbf{\begin{tabular}[c]{@{}c@{}}Actual \\ Value\end{tabular}}}} & \multicolumn{1}{c|}{1} & \multicolumn{1}{c|}{\begin{tabular}[c]{@{}c@{}}True\\ Positive (TP)\end{tabular}}   & \multicolumn{1}{c|}{\begin{tabular}[c]{@{}c@{}}False \\ Negative (FN)\end{tabular}} \\ \cline{2-4} 
\multicolumn{1}{c|}{}                                                                                  & \multicolumn{1}{c|}{0} & \multicolumn{1}{c|}{\begin{tabular}[c]{@{}c@{}}False \\ Positive (FP)\end{tabular}} & \multicolumn{1}{c|}{\begin{tabular}[c]{@{}c@{}}True\\ Negative (TN)\end{tabular}}   \\ \cline{2-4} 
\end{tabular}

\caption{Confusion matrix for pixel classification. Water pixels are represented by a value of 1 and land pixels are represented by a value of 0. }
\label{tab:confusion_matrix}
\end{table}

\begin{equation}\label{eq:accuracy} Accuracy = \frac{TP + TN}{TP + FP + TN + FN}  \end{equation}
\begin{equation}\label{eq:precision} Precision = \frac{TP}{TP + FP}  \end{equation}
\begin{equation}\label{eq:recall} Recall = \frac{TP}{TP + FN}  \end{equation}
\begin{equation}\label{eq:f1} F1 = \frac{2*Precision*Recall}{Precision + Recall}  \end{equation}

When calculating these metrics all pixels in an image are considered. As mentioned we expected the pixels close to the coastline to be harder to classify. By taking the general performance, these metrics may overestimate the performance of models in these regions. Hence, we also consider variations of these metrics where only the pixels within 10 pixels of a coastline pixel are considered. The coastline pixels are determined using the process detailed in the next section. 

\subsubsection*{Figure of Merit}

\begin{figure}[ht]
\centering
\includegraphics[width=0.80\textwidth]{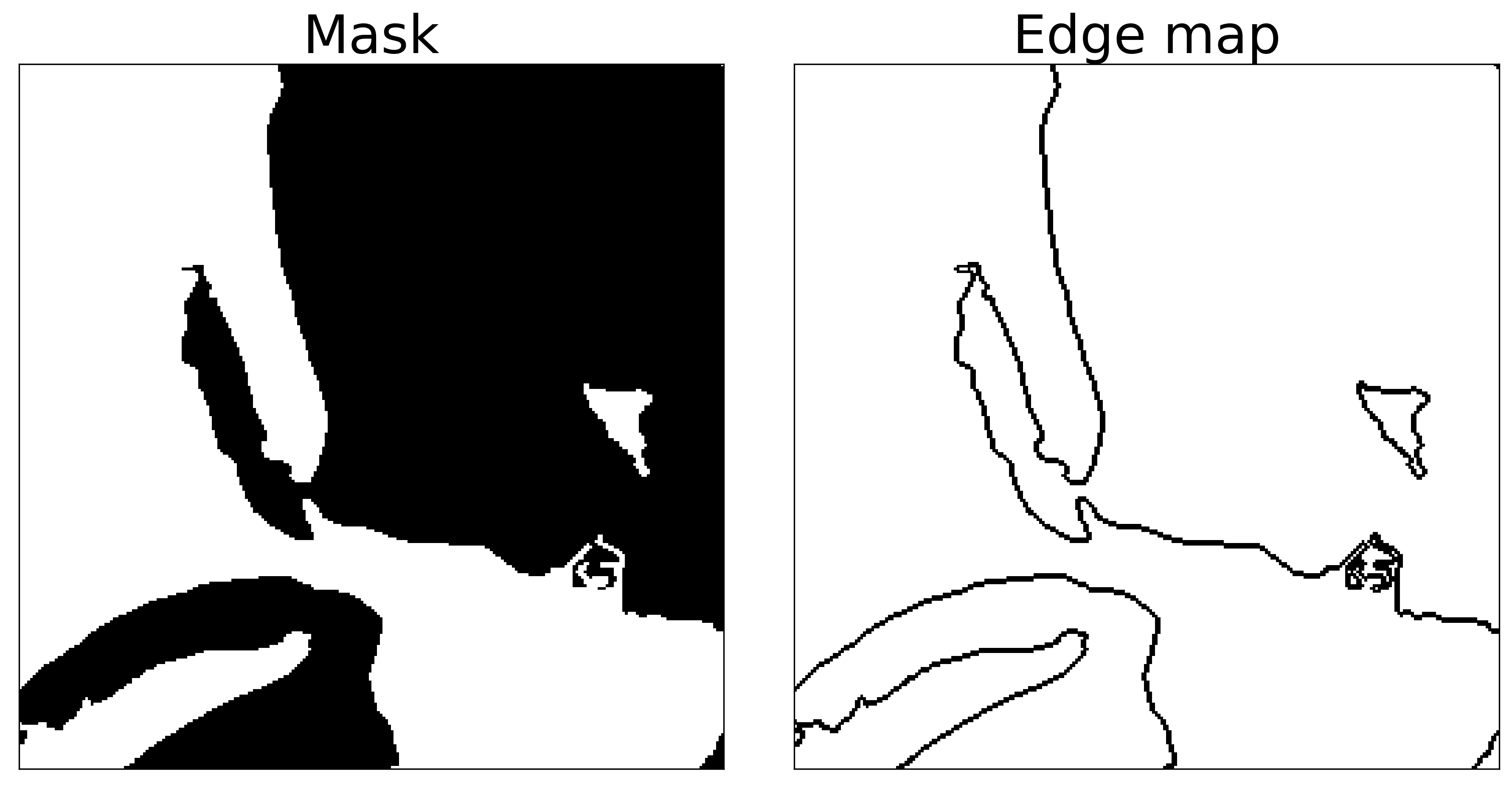}
\caption{Example of an edge map created using gradients of a mask. }
\label{fig:edge}
\end{figure}

We use Figure of Merit (FOM) as another approach for assessing the accuracy of the coastline. Previous experiments have shown this to be an effective metric for evaluating coastline edge detection problems~\cite{osullivan2023metric}. This metric is used for evaluating edge detection algorithms. Hence, as seen in Figure~\ref{fig:edge}, we must first create edge maps for the test masks and predictions. To do this we first calculate the gradient of each pixel. Pixels with a gradient that is not equal to 0 is labelled as an edge pixel. 

FOM is calculated using equation~\ref{eq:fom}. $N_G$ is the number of actual edge pixels, $N_E$ is the number of the detected edge pixels, $\alpha$ is the scaling constant, and $d(k)$ is the minimum distance between the detected edge pixel and an actual edge pixel~\cite{tariq2021quality}. In the context of our problem, FOM captures the average distance of the predicted coastline from the ground truth coastline. 

\vspace{0.1cm}
\begin{equation}\label{eq:fom} FOM(E,G) = \frac{1}{max(N_E, N_G)} \sum_{k=1}^{N_E} \frac{1}{1+\alpha d^2(k)}  \end{equation}
\vspace{0.1cm}

\subsection{Interpretability Metric}
To interpret the deep learning models, we use a permutation feature importance approach~\cite{osullivan2023interpreting}. This involves permuting the pixels in each band of the 100 test instances. The permutation score for each band is the original model accuracy less the accuracy when that band is permuted. Large values for this band suggest that the band was important to a model's predictions. This allows us to test the assumption that deep learning models benefit from using multiple bands as input. Understanding, which spectral bands are most important to predictions also builds trust in model predictions. This is because we can relate the results to previous research on spectral indices. A final benefit is that it will inform choices around future model development.

\section{Results \& Discussion}
\label{sec:results}
\subsection{Evaluation metrics}
Table~\ref{tab:metrics_overall} gives the evaluation metrics when all pixels are used in the calculations. For the deep learning approaches, U-NET had the highest accuracy of 95.0\%. This is 2.4 percentage points higher than XGBoost. This suggests that model performance is improved when pixel context can be used to classify each pixel. However, we see that the NDWI benchmark had better evaluation approaches in all metrics except recall. The average accuracy for NDWI was 2.2 percentage points higher than U-NET. FOM also indicates that NDWI was able to better approximate the coastline than the other approaches. 

\begin{table}[ht]
\caption{Evaluation metrics for the segmentation approaches applied to the LICS test set. The average of the evaluation metrics over 100 test images is given. }
\centering
\begin{tabular}{|c|c|c|c|c|c|c|}
\hline
\textbf{Method} & \textbf{Acc.} & \textbf{Prec.} & \textbf{Rec.} & \textbf{F1}      & \textbf{FOM}   \\ \hline
NDWI            & \textbf{0.972}    & \textbf{0.994}     & 0.946           & \textbf{0.967}      & \textbf{0.718} \\ \hline
XGBoost         & 0.926             & 0.990              & 0.842           & 0.897               & 0.440          \\ \hline
UNET            & 0.950             & 0.925              & \textbf{0.968}  & 0.941               & 0.546          \\ \hline
ATTUNET         & 0.947             & 0.960              & 0.919           & 0.927              & 0.556          \\ \hline
R2UNET          & 0.912             & 0.962              & 0.840           & 0.879               & 0.330           \\ \hline
\end{tabular}

\label{tab:metrics_overall}
\end{table}

Table~\ref{tab:metrics_coastline} gives the evaluation metrics when only the pixels within 10 pixels of a coastline edge are used in the calculations. Comparing the metrics to those in Table~\ref{tab:metrics_overall},  we see a decrease in all the values. This means that all methods had more difficulty predicting pixels close to the coastline than the pixels in general. Additionally, we now see larger differences between the methods. The average accuracy for NDWI is 10.2 percentage points higher than U-NET. This tells us that the improvement in NDWI over U-NET seen in Table~\ref{tab:metrics_coastline} comes primarily from more accurate predictions around the coastline. 

\begin{figure}[h]
\centering
\subfloat{
  \includegraphics[width=0.98\textwidth]{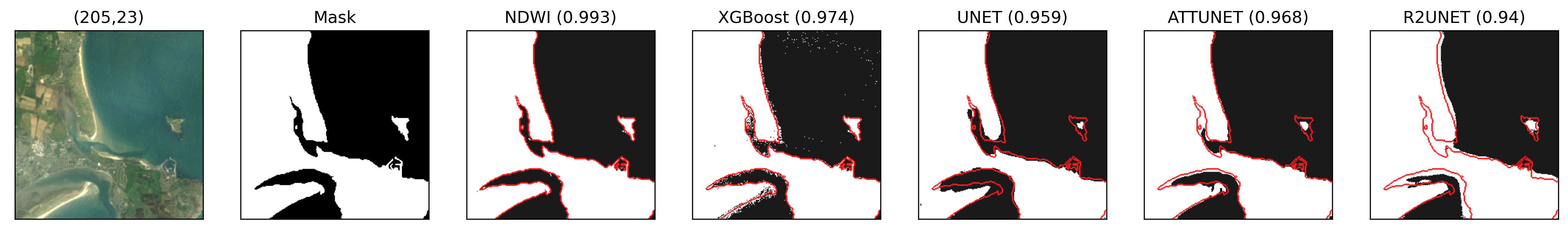}
  \label{fig:sub1}
} \vspace*{-1em}

\subfloat{
  \includegraphics[width=0.98\textwidth]{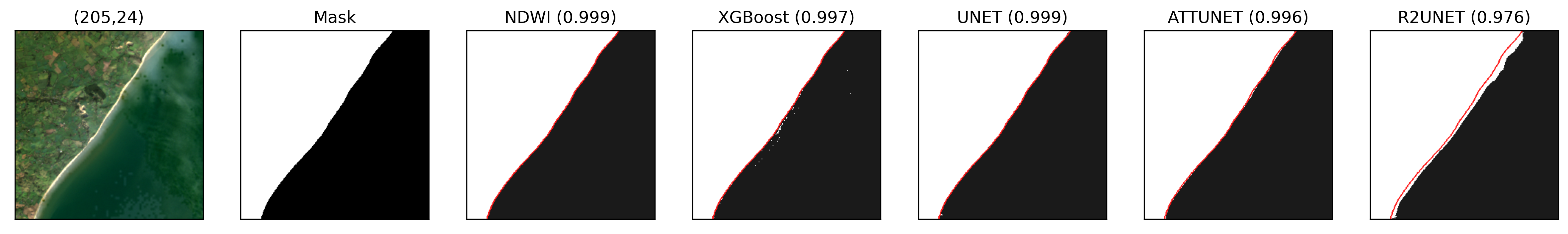}
  \label{fig:sub2}
} \vspace*{-1em}







\subfloat{
   \includegraphics[width=0.98\textwidth]{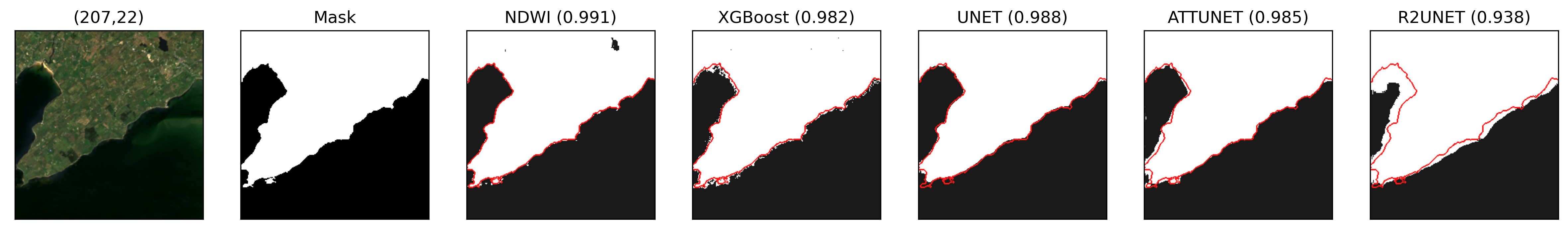}

} \vspace*{-1em}






\subfloat{
   \includegraphics[width=0.98\textwidth]{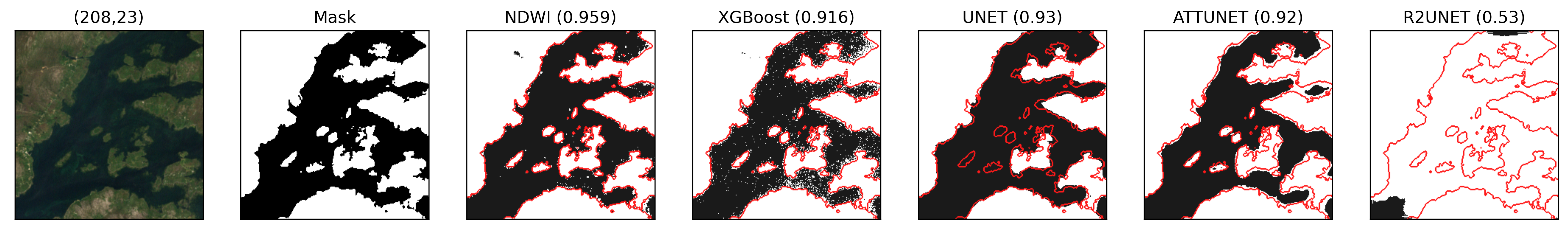}

} \vspace*{-1em}

\subfloat{
   \includegraphics[width=0.98\textwidth]{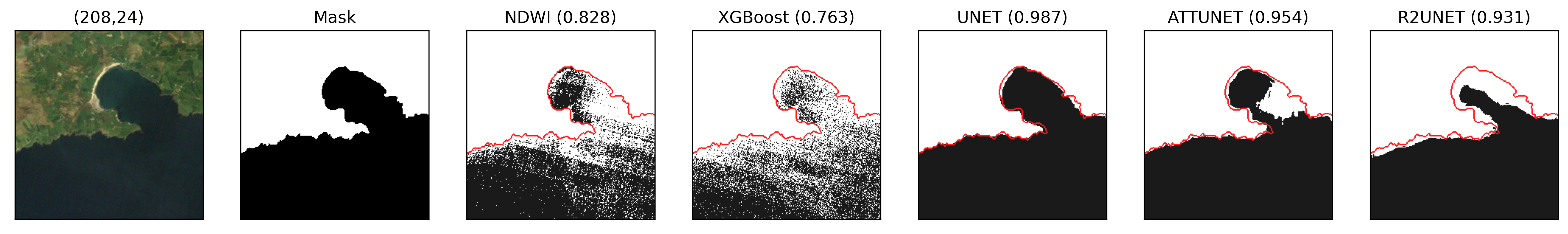}
} \vspace*{-1em}

\caption{Examples of predicted masks. Each row gives a random test image from a tile. The tile is given above the RGB visualisation in the first column. The second column gives the test mask for that image. The remaining 5 columns give the predicted mask for the 5 different approaches. The number next to the approach name gives the accuracy for that prediction when compared to the mask. The red line overlaying the predictions gives the edge of the coastline given in the mask. Examples for the remaining tiles can be found in the appendix.}

\label{fig:mask_examples}
\end{figure}

\begin{table}[ht]
\caption{Evaluation metrics within 10 pixels of the coastline. The average of the evaluation metrics over 100 test images is given.}
\centering
\label{tab:metrics_coastline}
\begin{tabular}{|c|c|c|c|c|}
\hline
\textbf{Method} & \textbf{Accuracy} & \textbf{Precision} & \textbf{Recall} & \textbf{F1}    \\ \hline
NDWI            & \textbf{0.938}    & \textbf{0.983}     & 0.891           & \textbf{0.927} \\ \hline
XGBoost         & 0.840             & 0.968              & 0.701           & 0.792          \\ \hline
UNET            & 0.836             & 0.822              & \textbf{0.905}  & 0.848          \\ \hline
ATTUNET         & 0.859             & 0.899              & 0.811          & 0.833          \\ \hline
R2UNET          & 0.720               & 0.895              & 0.527          & 0.618          \\ \hline
\end{tabular}
\end{table}
A visual analysis of Figure~\ref{fig:mask_examples} supports these results. We see that the U-NET predicts masks that tend to either under or over-estimate the coastline. In comparison, NDWI accurately predicts the coastline for most instances but misclassified ocean pixels further away from land. This is seen in the images for tiles (205,23) and (208,24). XGBoost is impacted in a similar way. Both of these methods only consider the intensity of individual pixels and the misclassified pixels will likely have intensity values similar to land pixels. In comparison, the deep learning approaches do not tend to misclassify pixels in this way. This is likely a result of including pixel context in predictions.

\subsection{The Advantages and Disadvantages of the Annotation Process}

We should consider the above results with the dataset annotation process in mind. The test set was annotated to provide precise segmentation masks. However, there was no on-site evaluation to ensure their accuracy. Further bias can be introduced as the annotations were not cross-evaluated by other professionals. In other words, the ground truth was determined by only one of the paper's authors using a visual analysis of the satellite images and Google Earth images of the same location.  

\begin{figure}[ht]
\centering
\includegraphics[width=0.99\textwidth]{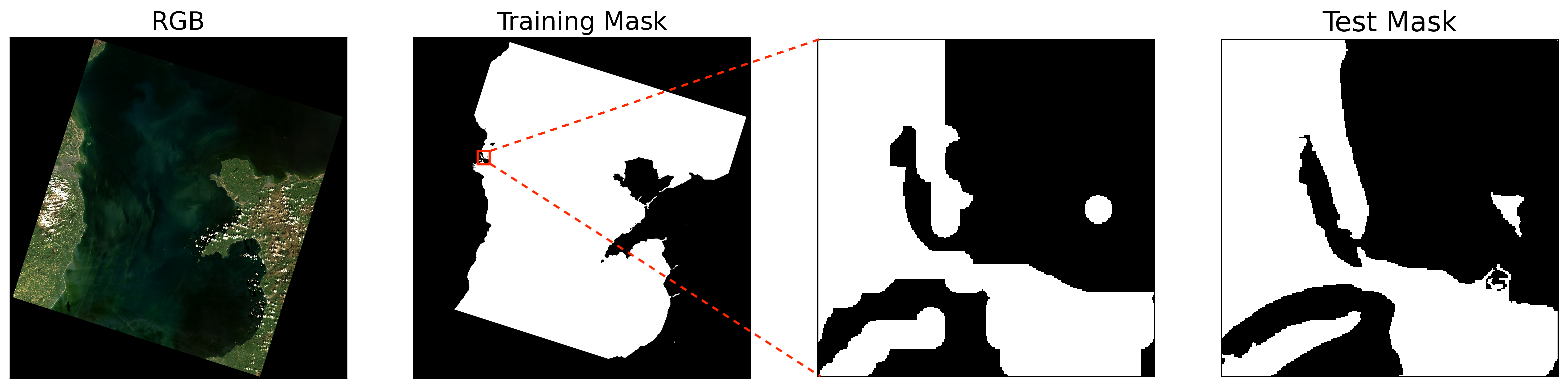}
\caption{Example of the output from the training and test annotation process. The training masks are obtained by cropping a mask drawn on the entire Landsat scene. When we zoom in on this mask, you can see there are incorrectly labelled pixels. In comparison, the test masks are obtained by annotating only the test location with more precision. }
\label{fig:train_annotation}
\end{figure}

A similar process was used for the training dataset. However, to ensure a reasonable amount of time was required to develop this dataset, the annotation process produced less precise segmentation masks. As you can see in Figure~\ref{fig:train_annotation}, this means there are incorrectly labelled pixels used to train the machine learning approaches. This helps explain the lower accuracy for these approaches compared to NDWI. Additionally, the way Landsat tiles were chosen may also limit the model's robustness as we have relied on relatively cloudless images. A final limitation is that the method for developing the training data was labor-intensive. In comparison, the NDWI benchmark requires no training data. 

There are still some noticeable benefits to the deep learning approach. Firstly, although we have shown a rough training mask for a testing location in Figure~\ref{fig:train_annotation}, we ensured that we would not include these locations in the training set. This is to ensure the results indicate how well the model can generalise to unseen locations. Secondly, the U-NET model, though initially trained on this dataset, can be fine-tuned and improved over time. Lastly, the dataset should still produce a model that is robust to other factors like solar altitudes, coastline types and time. We explore these factors in more depth using the NDWI, U-NET model and the accuracy metric.

\subsection{Accuracy by Coastal Type}
The LICS dataset was not designed specifically to analyse coastal types. Yet, through randomly selecting testing locations we can expect to capture variations in this factor. We can see this in Figure~\ref{fig:mask_examples} where various coastal types are present. As mentioned the testing location of the tiles, where further classified as having either a rocky or sandy coastline type. Another characteristic is the shape of the coastlines. That is some tiles are jagged and others more uniform. 

Table~\ref{tab:metrics_tile} gives the average accuracy from each tile. U-NET had the lowest accuracy for tile (208,23) which visually is the least uniform. In contrast, U-NET had the highest accuracy for tiles (208,24) and (205,24) which are relatively uniform. This suggests the model is not robust to variations in this coastline characteristic. As mentioned, the west coast of Ireland is more exposed to swell leading to more jagged coastlines. As a result, we may expect the model to perform worse in these regions.

\begin{table}[ht]
\caption{Average accuracy by tile for NDWI and U-NET. N gives the number of test images for each tile. }
\label{tab:metrics_tile}
\centering
\begin{tabular}{|c|c|c|c|}
\hline
\textbf{Tile} & \textbf{N} & \textbf{NDWI} & \textbf{UNET} \\ \hline
(205,23)      & 11         & 0.990         & 0.930         \\ \hline
(205,24)      & 20         & 0.966         & 0.985         \\ \hline
(206,22)      & 9          & 0.982         & 0.978         \\ \hline
(206,23)      & 6          & 0.996         & 0.982         \\ \hline
(206,24)      & 10         & 0.976         & 0.920         \\ \hline
(207,22)      & 9          & 0.934         & 0.966         \\ \hline
(207,23)      & 10         & 0.991         & 0.876         \\ \hline
(207,24)      & 7          & 0.994         & 0.964         \\ \hline
(208,22)      & 6          & 0.959         & 0.980         \\ \hline
(208,23)      & 6          & 0.955         & 0.866         \\ \hline
(208,24)      & 6          & 0.943         & 0.989         \\ \hline
\end{tabular}

\end{table}

Table~\ref{tab:metrics_type} gives the average accuracy of the tiles in each of these groups. For NDWI, the accuracy for the sandy coastlines is 2.5 percentage points higher than for rocky coastlines. This suggests it is potentially harder to segment rocky coastlines. However, we see the opposite for U-Net with a smaller difference of 1.1 percentage points. Considering this, in light of Table~\ref{tab:metrics_tile} it seems as if the coastline type does not influence model performance as much as its shape. 

\begin{table}[ht]
\caption{Average accuracy by coastline type for the NDWI and UNET approaches. The 11 tiles in the test set have been classified as either sandy or rocky coastlines. N gives the number of test images in each category. }
\label{tab:metrics_type}
\centering
\begin{tabular}{|c|c|c|c|}
\hline
\textbf{Type} & \textbf{N} & \textbf{NDWI} & \textbf{UNET} \\ \hline
sandy               & 67         & 0.980         & 0.947         \\ \hline
rocky            & 33         & 0.955         & 0.958  
\\ \hline
\end{tabular}
\end{table}

\subsection{Accuracy through Time}

In Table~\ref{tab:metrics_decade} we can see some variation when comparing accuracy for U-Net by decade. Specifically, the difference between the best (2010) and worst (2020) performing decades was 2.3 percentage points. However, we must consider potential confounding between the tiles and years. In Table~\ref{tab:per_tiles}, we see the percentage of test instances that come from the tiles for each decade. 18\% of the test instances for 2020 came from tile (208,23). Whereas this figure was 0\% for 2010. Considering that this was the tile with the lowest accuracy, it would partially explain the lower accuracy for 2020 in general.  In other words, variation in accuracy by decade is not due to inherent characteristics of scenes in that decade but the non-uniform distribution of tiles across the decades. 

\begin{table}[ht]
\caption{Average accuracy by decade for the NDWI and UNET approaches. N gives the number of test images for each decade. }
\label{tab:metrics_decade}
\centering
\begin{tabular}{|c|c|c|c|}
\hline
\textbf{Decade} & \textbf{N} & \textbf{NDWI} & \textbf{UNET} \\ \hline
1980            & 15         & 0.985         & 0.951         \\ \hline
1990            & 27         & 0.961         & 0.951        \\ \hline
2000            & 24         & 0.971         & 0.946         \\ \hline
2010            & 23         & 0.969         & 0.960         \\ \hline
2020            & 11         & 0.989         & 0.937        \\ \hline
\end{tabular}
\end{table}

\begin{table}[ht]
\caption{Percentage of test images that come from each tile in each decade.}
\label{tab:per_tiles}
\centering
\begin{tabular}{|c|c|c|c|c|c|}
\hline
\textbf{Tile} & \textbf{1980} & \textbf{1990} & \textbf{2000} & \textbf{2010} & \textbf{2020} \\ \hline
(205,23)      & 7             & 15            & 8             & 13            & 9             \\ \hline
(205,24)      & 13            & 26            & 13            & 22            & 27            \\ \hline
(206,22)      & 7             & 7             & 4             & 17            & 9             \\ \hline
(206,23)      & 13            & 11            & 4             & 0             & 0             \\ \hline
(206,24)      & 7             & 7             & 13            & 17            & 0             \\ \hline
(207,22)      & 20            & 11            & 4             & 9             & 0             \\ \hline
(207,23)      & 13            & 11            & 8             & 9             & 9             \\ \hline
(207,24)      & 20            & 0             & 4             & 9             & 9             \\ \hline
(208,22)      & 0             & 0             & 25            & 0             & 0             \\ \hline
(208,23)      & 0             & 4             & 13            & 0             & 18            \\ \hline
(208,24)      & 0             & 7             & 4             & 4             & 18            \\ \hline
\end{tabular}
\end{table}

\subsection{Solar Altitude}

In Table~\ref{tab:metrics_altitude}, we see the performance across the different altitude categories. For U-NET the difference between the best and worst altitudes is 1.5 percentage points. This figure is 2.1 percentage points for NDWI. In contrast to previous research the spectral indices performed better for lower altitudes. Even so, the results suggest that solar altitude can have an impact on the performance of spectral indices. In comparison, the performance of U-NET is more uniform. This suggests that solar altitude does not play a significant role in the ability of the model to perform accurate segmentation. 

\begin{table}[ht]
\caption{Average accuracy by altitude category for the NDWI and UNET approaches. N gives the number of test images in each category. }
\label{tab:metrics_altitude}
\centering
\begin{tabular}{|c|c|c|c|}
\hline
\textbf{Altitude} & \textbf{N} & \textbf{NDWI} & \textbf{UNET} \\ \hline
low               & 34         & 0.984         & 0.951         \\ \hline
medium            & 34         & 0.963         & 0.943         \\ \hline
high              & 32         & 0.969         & 0.958         \\ \hline
\end{tabular}
\end{table}

These results show promise that a robust deep-learning model can be built. The model did have lower performance for some coastline shapes. We believe this can be addressed through more accurate training annotations. At the same time, the model produced similar results for different decades, coastal types and solar altitude which is a proxy for time of year. Hence, the results show that a deep learning model can be used for inference for any scene in Ireland, during any time of the year, provided that scene is not cloudy.  


\subsection{Permutation Band Importance}

\begin{figure}[h]
\centering
\includegraphics[width=0.80\textwidth]{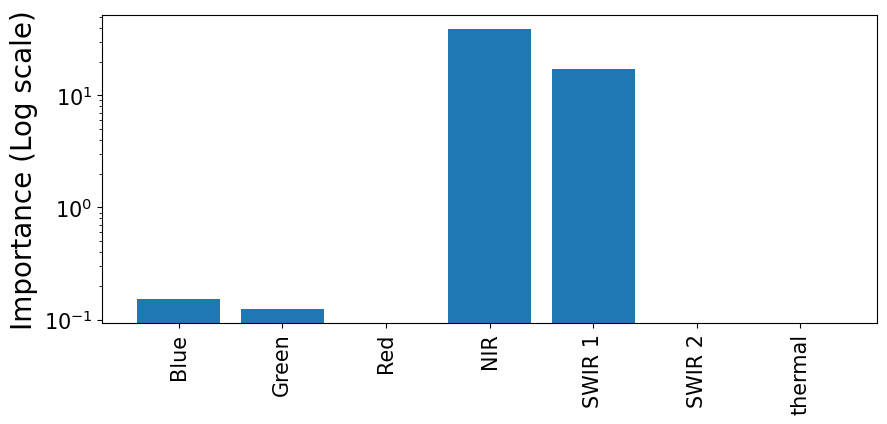}
\caption{Permutation importance scores from each band used in the U-NET model. Accuracy is decreased by 38.96 and 17.17 percentage points when the NIR and SWIR 1 bands are permuted.  }
\label{fig:importance}
\end{figure}

The visual analysis of the segmentation predictions in Figure~\ref{fig:mask_examples} suggests that the deep learning models benefit from using pixel context. This is a commonly stated benefit of deep learning models over spectral indices. Another stated benefit is they can use all available spectral bands as input. Looking at Figure~\ref{fig:importance}, we can see that for the U-NET approach this benefit may be overstated. The largest permutation scores are 38.96\% and 17.17\% for the NIR and SWIR 1 bands respectively. The blue and green bands had small positive scores of 0.15\% and 0.12\% respectively. The remaining scores were small negative values. This suggests that only the NIR and SWIR 1 bands are having a significant impact of model predictions. 

The NIR band is recognised as an important spectral band for water body segmentation. It is used in the NDWI indices included in this paper. it is also used in the calculation for the Automated Water Extraction Index with Shadows Elimination (AWEIsh)~\cite{feyisa2014automated} and Water Index 2015 (WI2015)~\cite{fisher2016comparing}. Likewise, the SWIR 1 band is used to calculate AWEIsh and WI2015 as well as a modified version of the NDWI (MNDWI)~\cite{xu2006modification}. Ultimately, through the process of training, the model has identified bands that have been used in spectral indices.

\section{Conclusion \& Future Work}
\label{sec:conclusion}
We presented LICS, the first Irish coastline segmentation dataset for deep learning. It was created using Landsat scenes from 1984 to 2023 and includes 30,000 training instances and 100 test instances. We benchmarked the performance on this dataset using various segmentation approaches including an NDWI threshold, XGBoost and the U-NET deep learning architecture. For the benefit of the community, both the dataset and code for these experiments are made freely available. 

When developing the LICS dataset, we aimed to capture variation in factors, inherent to the Irish coast, that were expected to impact model performance. These include the year and month of the scene, coastline types and solar altitude. Initial results suggest that it is possible to build a deep learning model that is robust to changes in these factors. This means that such a model can output accurate segmentation for any Landsat scene of the Irish coastline. This will enable accurate inference and further coastal monitoring efforts. 

We explored assumptions around the benefits of deep learning approaches, such as U-NET, over other segmentation methods. These are that U-NET can use pixel context and all available spectral bands to make predictions. A visual analysis of U-NET predictions verified the first benefit. Interpreting the U-NET showed that the second benefit is not as influential. Results suggested only the NIR and SWIR 1 bands were used to make predictions. Future models can take advantage of this result. By using only the two bands as input, we can reduce model complexity and training time whilst having no negative effect on model performance. 

It is important to not overstate the performance of the deep learning approaches. The results do not show an improvement over traditional spectral indices. U-NET was the best-performing model with an average accuracy of 95.0\%. This is compared to 97.2\% when using NDWI. However, a visual analysis showed promise for the deep learning approaches. U-NET tended to misclassify pixels close to the coastline. This is likely a result of the annotation process for the training set. It produced rough masks where the pixels close to the coastline were most likely to be incorrectly labelled. 

We believe that the deep learning approach can significantly outperform the spectral indices given more accurate training data. Future research will focus on developing a modelling process that will create accurate annotations while limiting the amount of time required to label training data. This will likely involve semi-supervised methods used to annotate a large number of training instances as well as a smaller manually annotated dataset. This will enable a transfer learning approach where an initial model, trained on the semi-supervised dataset, can be fine-tuned on the manually annotated dataset. 

When pursuing this goal we must consider the purpose of the model. The dataset was developed based on the instantaneous coastline definition. This fundamentally limits a model's ability to monitor erosion and other coastline changes. Additionally, the 30m resolution of Landsat scenes means that only changes over relatively long periods can be observed. Future research will focus on alternative definitions such as the high water mark, vegetation line and dune volume and use higher resolution sources such as sentinel-2 satellite imagery. Variations to the instantaneous coastline definition will also be considered such as including a third category for mixed pixels. When doing all of this, we will explore how the LICS dataset can be leveraged using fine-tuning approaches. 

\section*{Acknowledgments}
This research was conducted with the financial support of Science Foundation Ireland under Grant Agreement No.\ 13/RC/2106\_P2 at the ADAPT SFI Research Centre at University College Dublin. The ADAPT Centre for Digital Content Technology is partially supported by the SFI Research Centres Programme (Grant 13/RC/2106\_P2) and is co-funded under the European Regional Development Fund. This publication has emanated from research conducted with the financial support of Science Foundation Ireland under Grant number 18/CRT/6183. For the purpose of Open Access, the author has applied a CC BY public copyright licence to any Author Accepted Manuscript version arising from this submission. 

\pagebreak
\appendix
\section{Additional Figures}
\label{sec:sample: appendix}
\begin{figure}[h]
\centering
\subfloat{\includegraphics[width=0.98\textwidth]{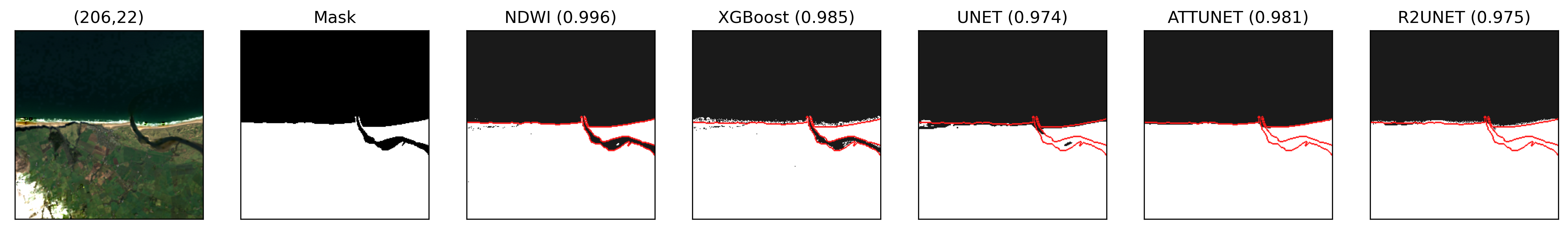}} \vspace*{-1em}

\subfloat{\includegraphics[width=0.98\textwidth]{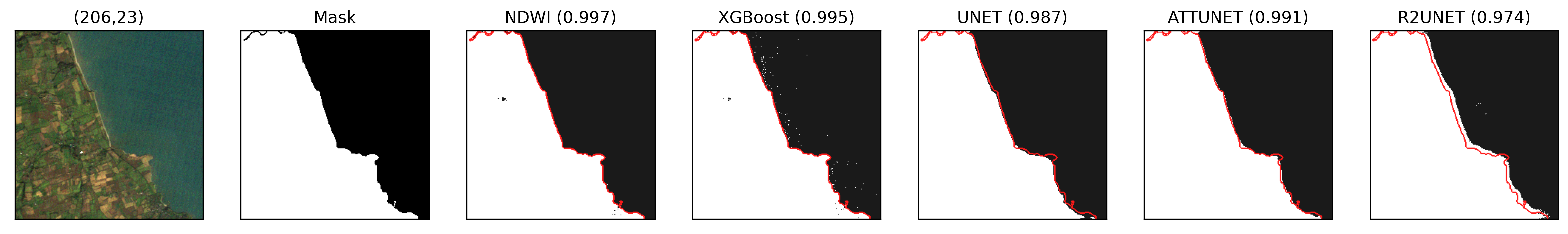}} \vspace*{-1em}

\subfloat{\includegraphics[width=0.98\textwidth]{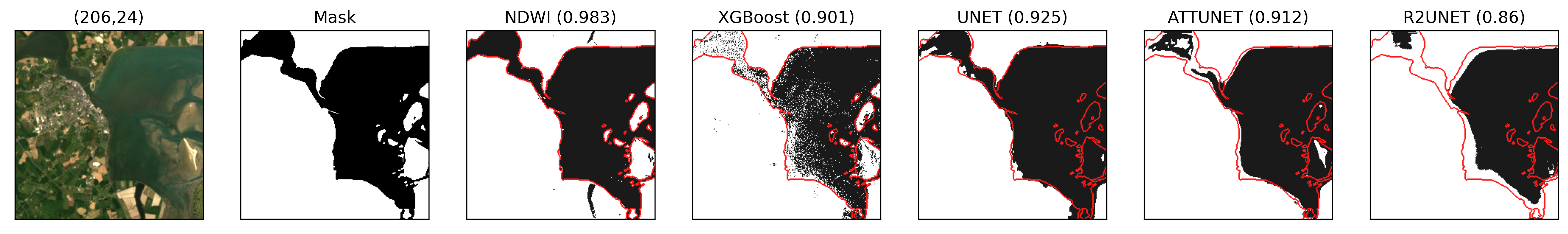}} \vspace*{-1em}

\subfloat{\includegraphics[width=0.98\textwidth]{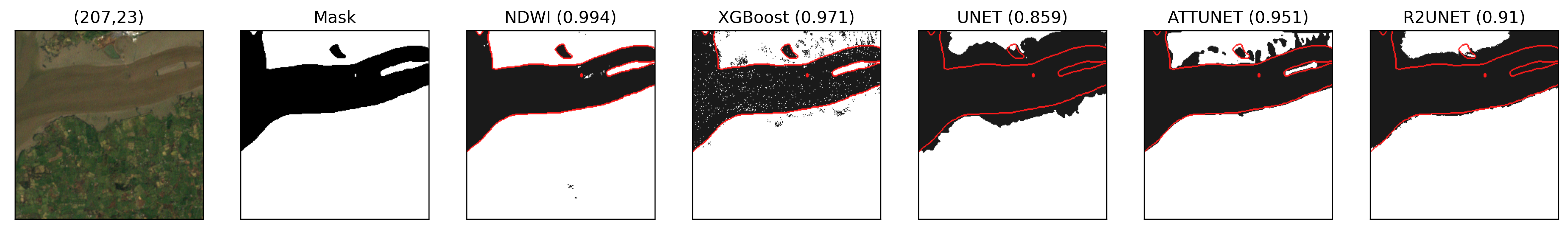}} \vspace*{-1em}

\subfloat{\includegraphics[width=0.98\textwidth]{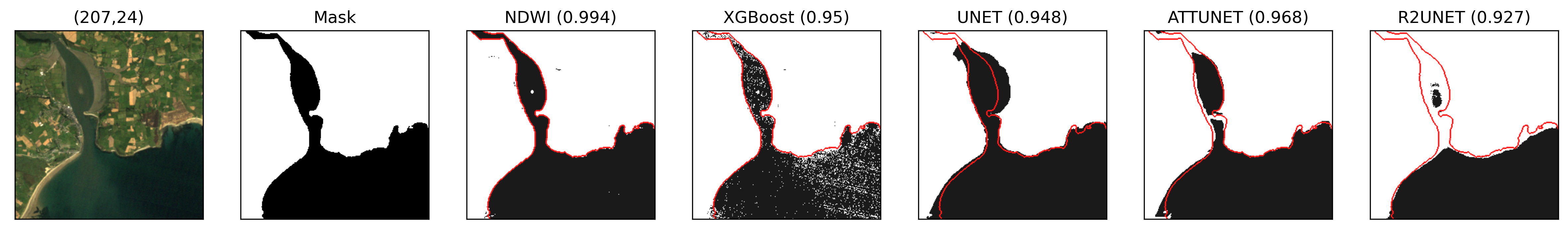}} \vspace*{-1em}

\subfloat{\includegraphics[width=0.98\textwidth]{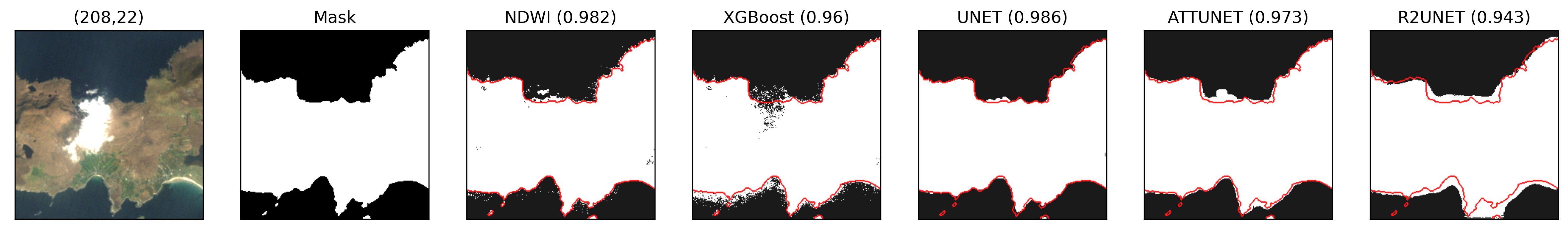}} \vspace*{-1em}

\caption{Additional examples of predicted masks from the segmentation approaches.}

\label{fig:mask_examples_2}
\end{figure}

\bibliography{references}

\balance

\end{document}